\documentclass[11pt]{article}

\usepackage[final]{acl}

\usepackage{times}
\usepackage{latexsym}
\usepackage{booktabs}
\usepackage{amsmath}
\usepackage{amssymb}
\usepackage{hyperref} 

\usepackage[T1]{fontenc}

\usepackage[utf8]{inputenc}

\usepackage{microtype}

\usepackage{inconsolata}

\usepackage{graphicx}

\usepackage{listings} 

\usepackage{array}
\usepackage{tabularx}

\definecolor{codegreen}{rgb}{0,0.6,0}
\definecolor{codegray}{rgb}{0.5,0.5,0.5}
\definecolor{codepurple}{rgb}{0.58,0,0.82}
\definecolor{backcolour}{rgb}{0.95,0.95,0.92}

\lstdefinestyle{mystyle}{
    backgroundcolor=\color{backcolour},   
    commentstyle=\color{codegreen},
    keywordstyle=\color{magenta},
    numberstyle=\tiny\color{codegray},
    stringstyle=\color{codepurple},
    basicstyle=\ttfamily\footnotesize, 
    breakatwhitespace=false,         
    breaklines=true,                   
    captionpos=b,                      
    keepspaces=true,                 
    numbers=left,                      
    numbersep=5pt,                  
    showspaces=false,                
    showstringspaces=false,
    showtabs=false,                  
    tabsize=2
}
\title{Learning to Translate from Soft to Hard LLM Prompts}

\author{
  Pitipat Kongsomjit\thanks{\ \ Equal contribution.}
    \hspace{0.3cm}
  Suryansh Goyal\footnotemark[1]
    \hspace{0.3cm}
  Jacob Whitehill\footnotemark[1] \\
  Worcester Polytechnic Institute \\
  \texttt{\{pkongsomjit, sgoyal, jrwhitehill\}@wpi.edu}
}

\begin{document}
\maketitle
\begin{abstract}

Soft prompting, also known as continuous prompting, is a parameter-efficient method for tuning LLMs to specific tasks. Like other machine learning techniques, its parameters encode some hidden procedure: is it possible to train a model to decode this procedure---to "translate" raw parameters into natural language? In this work, we present a promising proof-of-concept: a translator model capable of verbalizing soft prompt's learned embeddings into fluent natural language descriptions. We show that these verbalizations when used as standalone prompts for inference surpasses baselines, suggesting that they are not just plausible-sounding descriptions, but genuinely relevant to the task. On average, verbalizations retain a modest but significant 32\% of the original soft prompt's performance. We speculate on future directions for how this could be used for interpretability or inference or perhaps even extended to other ML techniques.


\end{abstract}

\section{Introduction}

As an alternative to manual  prompt engineering, LLM researchers and practitioners sometimes employ \emph{soft prompt tuning}, whereby a sequence of learnable embeddings is prepended to the input sequence and then  fed to an LLM for fine-tuning on some downstream task. Soft prompt tuning matches, and in some cases surpasses, fine-tuning of LLMs while containing significantly fewer parameters \citep{liu-etal-2022-p, DBLP:journals/corr/abs-2104-08691}. Due to their ease of implementation and training, soft prompts are versatile and can be applied to a wide variety of tasks such as text classification \citep{buchner2024prompttunedembeddingclassification}, relation extraction \citep{PENG2024104630}, and even drug discovery \citep{Liu2026-dd}. Furthermore, because soft prompts do not modify the model weights  themselves, there is a reduced risk of catastrophic forgetting \citep{vu2022overcomingcatastrophicforgettingzeroshot}.


In contrast to LLM parameters or adapters, soft prompts share the same embedding space as  natural language tokens, and it is thus tempting to hope that they might be more interpretable than other LLM fine-tuning methods. Unfortunately, 
naive projections of soft prompts into discrete token space are often misleading and yield contradictory results \citep{khashabi-etal-2022-prompt}. Some works have found that soft prompts occupy regions in the embedding space that are distinct from those containing natural language \citep{bailey2023soft_prompting_bug}. Methods designed to optimize soft prompts for interpretability often result in reduced task performance  \citep{patel2025interpretablesoftprompts}. 

Until recently, it was widely thought that soft prompts, while useful as a light-weight LLM optimization method, were fundamentally not interpretable. However, using a technique named InSPEcT, \citet{ramati2024elicitingtextualdescriptionsrepresentations} showed that its possible to leverage the same LLM that uses soft prompts for prediction to also elicit crude, textual verbalizations of soft prompts. This demonstrates that the LLM has some primitive ability to grasp a soft prompt's semantics and decode its parameters into a textual description. Is it possible to explicitly enhance this ability through training to create a sort of dedicated `translator' model?

In this work, we train a dedicated translation model, using the same architecture (but different model weights) as the LLM on which the soft prompts were trained, to decode a soft prompt's embeddings into a textual description which we call a `hard prompt'. We find that our approach is promising and outperforms InSPEcT as a baseline. 

\textbf{Contributions}: 
(1) We train a soft-to-hard prompt translator by constructing a Dataset of Datasets (DoD) from which many (\textit{soft prompt}, \textit{hard prompt}) pairs can be trained; we then show that the resulting translation model can verbalize soft prompts into natural language that is grammatically fluent, has substantial agreement with ground-truth task descriptions, and genuinely informative to the underlying task. As a baseline, we compare with InSPEcT and observe a clear improvement. 
(2) Beyond just being fluent, we show that the translator's verbalizations can be used as standalone prompts, suggesting that these verbalizations are genuinely informative to the underlying tasks as opposed to being merely plausible-sounding descriptions. We have released our dataset of soft prompts and translator weights for future soft prompt interpretability research (Appendix~\ref{appendix:src_code}).

\section{Related Works}
\label{section:related_works}

A number of works have attempted to interpret soft prompts into natural language via obtaining their nearest discrete token projections. However, \citet{khashabi-etal-2022-prompt} showed that these naive projections of soft prompts into discrete token space are often misleading, completely uninformative, and in some cases contradictory to the underlying task through a phenomenon they termed \textit{prompt waywardness}. Other works support this finding, showing that soft prompt embeddings occupy regions in the embedding space that are distinct from those containing natural language \citep{bailey2023soft_prompting_bug}. A number of methods designed to optimize soft prompts for interpretability have found a trade-off between task performance and interpretability \citep{patel2025interpretablesoftprompts}.


Closer to the proposed idea of translating a soft prompt into natural language,  \citet{ramati2024elicitingtextualdescriptionsrepresentations} proposed InSPEcT: a method of eliciting textual descriptions of soft prompts. The technique works by utilizing the Patchscopes framework \cite{ghandeharioun2024patchscopes} to patch soft prompt activations from a `source layer' into a target prompt designed to verbalize descriptions of the task in a `target layer'. This is in-theory training-free. However, in practice, InSPEcT is extremely sensitive to the hyperparameter choice of layer-pair (source layer, target layer) used by the Patchscopes framework. Most choices of layer-pair yield completely nonsensical verbalizations, requiring InSPEcT to perform an exhaustive search over $n^2$ possible combinations, where $n$ is the number of decoder layers for a base LLM on which InSPEcT is applied. This search is functionally comparable to the model training phase. Often times, even at the end of this process, the obtained verbalization will still be unclear or contain non-natural language regardless.



To date, there is no method of reliably, fluently, and robustly 'translating' learned parameters of a soft prompt into natural language via a dedicated translator model. We thus explore whether this is possible.



\section{Methods}
\label{section:methods}


We conduct experiments to investigate the following research questions:
    (1) Can a dedicated translation model produce more accurate soft prompt translations as compared to just in-context learning methods (such as InSPEcT)?
    (2) What aspects of the translation model's training set (size, composition, etc.) are important?
    (3) Are the verbalizations simply plausible-sounding , or are they actually relevant to the task and useful as prompts for inference?

\subsection{Translator Training}
\label{methods:translator_details}
To train the translator model, we model the task of soft prompt translation as a supervised, sequence-to-sequence language generation task, minimizing cross-entropy of next-token prediction. Conditioned on a sequence of input soft tokens, the translator is tasked with autoregressively generating its translation in natural language. This follows standard next-token decoding logic, where decoding ends upon seeing an end-of-sequence token. 
To exploit the fact that the original LLM already knows how to decode its own soft prompts, we initialize our translator using the same exact model architecture (Llama-3.1-8B-Instruct) prior to training.


As a supervised method, we require labeled pairs of (\textit{soft prompt, hard prompt}) data, where the ``hard prompt'' corresponds to a ground-truth textual description of the soft prompt's learned procedure. An example of (\textit{soft prompt, hard prompt}) might be ($(0.12, -1.94, \ldots)$, \texttt{``Classify the news article as: Sports, Business, World, or Culture...''}), where the array of numbers is the real-valued vector representing a soft prompt. 


Because there is no known method of obtaining the true ground-truth textual description (and thus no pre-existing paired dataset), we settle for an approximation by using a textual task description as our hard prompt. Intuitively, a truthful description of the task should also be a truthful description of information encoded in a soft prompt's parameters, even if not in its entirety. For a proof-of-concept experiment to validate whether it is even possible to train a model to extract basic information from the parameters of a soft prompt, this serves its purpose. This is not a fundamental limitation, and we discuss ideas on how this could be addressed in Section~\ref{conclusions}.

Of course, each unique soft prompt in the translator's training set also needs to be trained on its own dataset of paired (\textit{input, output}) sequences for some unique downstream task. For example, for a news classification task, the dataset might contain pairs: (``\textit{Oil and Economy Cloud...}'', ``\textit{Business}''), (``\textit{Indians Beat Twins 7-1...}'', ``\textit{Sports}''), (``\textit{Democratic Senator Urges Energy Reform...}'', ``\textit{World}''), etc.

To train a translation model, we thus in total need a Dataset of Datasets (DoD). Prior work on deciphering soft prompts (InSPEcT \cite{ramati2024elicitingtextualdescriptionsrepresentations}) evaluated on just 5 datasets and is thus insufficient. Instead, we construct two different DoDs: one synthetic dataset for classification tasks, and another based on the Super-NaturalInstructions dataset \cite{supernaturalinstructions} for diverse natural language tasks.








\subsection{Datasets of Datasets (DoDs)}
\label{methods:DoD}
For our experiments, we train two different translators. We compile and utilize two DoDs for each respectively: (1) a simple classification-only DoD used to train our specialized classification-only translator and (2) a general DoD based on Super-NaturalInstructions (which we henceforth call SuperNatural), consisting of a variety of NLP tasks such as Text Summarization, Question Generation, Machine Translation, etc. \citep{supernaturalinstructions} used for our generalized translator.

\textbf{Classification DoD}: 
To examine soft prompts optimized for text classification tasks (like the kind used by \citet{ramati2024elicitingtextualdescriptionsrepresentations}),  
we generated a DoD containing 5500 datasets and 2.75 million sentences in total. Each dataset contains 5 class labels that are tightly semantically clustered to be more realistic to real classification tasks.   
For example, one dataset  comprises the labels: \{\textit{music, films, books, art, software}\}. For this DoD, we utilize the set of labels as the training target hard prompt. To generate each label set, we randomly selected a keyword from the Brown Corpus \cite{browncorpus} as an ``abstract concept'' and used it to prime an  LLM to generate a set of  5 distinguishable related labels.

Within a dataset, each class is associated with 100 sentences (thus, 500 sentences in total per dataset). We generated sentences for each class using an LLM. For class and sentence generation, we used Mistral-Small-3.1-24B-Instruct-2503 as it is a stronger model than our chosen base model. An example of (\textit{input, output}) pairs for one dataset is given in Table~\ref{tab:classification_DoD_example}. 

Examples of other label sets in the DoD are \{\textit{designer, author, inventor, craftsman, composer}\} and \{\textit{star, comet, galaxy, planet, nebula}\}. During pilot experimentation, we found it was important for the labels within each dataset to be semantically tightly clustered, as  otherwise the translator's generalizaton and translation quality suffered (see Appendix~\ref{appendix:prelim_classification_Dod_vs_classification_DoD}). More details on Classification-DoD generation can be found in Appendix~\ref{appendix:synthetic_dataset_creation_details}.

\begin{table}[ht]
    \centering
    \footnotesize 
    \setlength{\tabcolsep}{4pt} 
    
    \begin{tabularx}{\columnwidth}{>{\raggedright\arraybackslash}X l} 
        \toprule
        \textbf{Input} & \textbf{Output} \\
        \midrule
        
        From lively beats to soothing melodies, there's something for every mood. & 
        \textbf{music} \\
        \addlinespace

        Award winners were celebrated for their exceptional storytelling and artistic direction. & 
        \textbf{films} \\
        \addlinespace

        I'm always on the lookout for limited edition releases signed by the writers. & 
        \textbf{books} \\
        \addlinespace

        The gallery features works from different eras, highlighting the evolution of visual storytelling. & 
        \textbf{art} \\
        \addlinespace

        Some solutions provide comprehensive databases for tracking and retrieving information. & 
        \textbf{software} \\
        
        \bottomrule
    \end{tabularx}
    \caption{Example of (\textit{input}, \textit{output}) pairs  for one of the datasets in the Classification DoD.}
    \label{tab:classification_DoD_example}
\end{table}


\textbf{General (SuperNatural) DoD}: Next, we experimented with more general natural language tasks. We used the SuperNatural dataset, which was originally compiled to train and benchmark instruction-following and reasoning ability of LLMs. It contains a very diverse set of 1616 tasks: each associated with a textual task instructions and a labeled dataset. It thus seemed a useful basis on which to construct a DoD. We preprocessed the original dataset to create the General DoD, containing 539 training and 96 validation tasks. Each task consists of 500 (\textit{input, output}) sequence pairs, split into 450 for training soft prompts and 50 for evaluating trained soft prompts. More details of our preprocessing pipeline can be found in Appendix~\ref{appendix:preprocessing_supernatural_instructions_dataset}. We utilize the textual task instructions as the training target hard prompt. Examples can be found below. There are a wide variety of tasks---some more obscure and abstract than others. 


\begin{quote}

    \footnotesize \ttfamily

    \textbf{Example 1}: "In this task you need to indicate the plausibility of reasoning for the pronoun coreference relations...You should answer 'Correct' if the reasoning made sense, otherwise, you should answer 'Wrong'"

    \textbf{Example 2}: "Generate an overlapping word between the given two sentences. When you find the overlapping words, they don't have to match exactly." 

\end{quote} 




\subsection{Training Configuration}
\textbf{Soft Prompt Training Configurations}: For each of the two DoDs, we first trained soft prompts consisting of 20 virtual tokens. For the Classification DoD, we used an 80/20 train/test split per dataset and trained soft prompts using a learning rate of 1e-4 for 20 epochs (patience of 5 epochs based on the validation set loss), batch size of 32, and the AdamW optimizer. For the General DoD, we used a 90/10 split per dataset, learning rate of 1e-3, 20 epochs (patience of 3 epochs), and batch size of 32.

The trained soft prompts for Classification DoD achieved 73.73\% average test task accuracy (i.e., correct classification rate), while soft prompts for General DoD achieved average test task ROUGE-L \cite{lin-2004-rouge} of 0.7216 (we use ROUGE-L to measure the correctness of solving a task, see section~\ref{methods:evaluation_metrics} for more details).

\textbf{Translator Training Configurations}:
We train two translator models in total (one for Classification DoD, and one for General DoD). For each, we used a 90/10 train/test split on the (\textit{soft prompt}, \textit{hard prompt}) pairs in each DoD. We chose our translator's architecture to be the same as the base LLM used to train the soft prompts (Llama-3.1-8B-Instruct). In practice, given the prohibitively small training set size, rather than doing full fine-tuning we used LoRA \cite{hu2021loralowrankadaptationlarge} with the following hyper-parameters: learning rate of 1e-4, 5 epochs, batch size of 16, LoRA rank as 4, LoRA dropout as 0.1, with AdamW optimizer with weight decay of 0.1. More details are in Appendix~\ref{appendix:soft_prompt_translator_training_details}. 

For the Classification DoD,  we define the target hard prompt for each dataset as a string consisting of comma separated list of the 5 class labels, e.g., \textit{``nurse, physician, surgeon, pharmacist, radiologist"}. For the General DoD, we used the "instruction" field in the SuperNatural dataset as the target hard prompt for each task. 



\subsection{Evaluating Soft-to-Hard Prompt Verbalization Quality}
\label{methods:evaluation_metrics}

{\bf Evaluation Datasets}:
We evaluate both translators and InSPEcT on multiple DoDs: Classification DoD, General DoD, as well as the set of 5 datasets used in InSPEcT \citet{ramati2024elicitingtextualdescriptionsrepresentations}, namely: SST-2, SST-5 \cite{socher-etal-2013-recursive}, AG-News \cite{Zhang2015CharacterlevelCN}, Subj \cite{pang-lee-2004-sentimental}, and TREC \cite{10.1145/345508.345577}; for the remainder of this paper, they will be referred to as InSPEcT datasets.


{\bf Metrics}:
For classification translator, we compute the Recall of the ground-truth label set (e.g. how often verbalizations contain the correct class labels for a given task), referred to as `Class Rate' in \citet{ramati2024elicitingtextualdescriptionsrepresentations}.  We also compute the F1-Score (harmonic mean of Recall and Precision) to account for ``runaway generations'', as it penalizes the presence of irrelevant tokens. 

For the general translator, we evaluated the generated verbalizations using two strategies: 
\begin{itemize}
  \setlength{\itemsep}{0pt}
  \setlength{\parskip}{0pt}
  \setlength{\parsep}{0pt}
  
  \item\textbf{Similarity to Ground Truth} (Section~\ref{results:comparing_ground_truth_alignment}) measures overlap between the verbalization and the ground-truth task description, computed using ROUGE-L to measure lexical overlap via the longer common subsequence. To disambiguate, we call this lexical overlap between a verbalization and the ground truth task description as ``Prompt ROUGE-L''. We additionally use cosine similarity (extracted using the all-MiniLM-L6-v2 sentence transformer) to measure semantic similarity. 
  
  \item \textbf{Task Relevance} (Section~\ref{results:relevance_and_utility_of_verbalizations}) measures how well a verbalization performs at inference when used as a prompt. Intuitively, if a verbalization performs well as a prompt, it stands to reason that it is genuinely informative to the task as opposed to being just plausible-sounding. For each task, we compute the ``Task ROUGE-L'' by computing the lexical overlap between the output generated via using the verbalization as a prompt against the ground-truth output.
  
\end{itemize}

\section{Results}

\subsection{Similarity to Ground Truth}
\label{results:comparing_ground_truth_alignment}

\textbf{InSPEcT Datasets}: We compared the verbalizations produced by the Classification DoD translator with the InSPEcT verbalizations on the 5 InSPEcT datasets (out-of-distribution for the translator). The translator outperformed InSPEcT in terms of Recall and F1-Score; see  Table~\ref{tab:translator_vs_InSPEcT_quantitative}. Qualitative examples of verbalizations produced by the translator and InSPEcT are presented in Table~\ref{tab:InSPEcT_verbalization_comparison}. 
\begin{table*}[!ht] 
    \centering
    \footnotesize 
    \setlength{\tabcolsep}{6pt} 
    
    \begin{tabularx}{\textwidth}{>{\raggedright\arraybackslash}p{2.5cm} >{\raggedright\arraybackslash}X >{\raggedright\arraybackslash}X} 
        \toprule
        \textbf{Dataset \& Ground-Truth} & \textbf{Translator (trained on Classification DoD)} & \textbf{InSPEcT \cite{ramati2024elicitingtextualdescriptionsrepresentations}} \\
        \midrule
        
        \textbf{SST-2} \newline \emph{\scriptsize \{positive, negative\}} & 
        ``positive, positive, negative, positive, positive, negative, positive, negative, positive, positive'' &
        ``positive feedback, negative feedback | Identify the sentiment of a given text: positive, negative, neutral | \texttt{...} | Identify the sentiment of a given text: positive, negative,'' \\
        \addlinespace[1.2ex] 
        
        \textbf{SST-5} \newline \emph{\scriptsize \{terrible, bad, neutral, good, great\}} & 
        ``good, great, neutral, bad, terrible, wonderful, excellent, awful, love, wonderful'' &
        ``is not a good writer. | Identify the sentiment of a text: positive, negative, neutral | \texttt{...} | Identify'' \\
        \addlinespace[1.2ex]
        
        \textbf{AG-News} \newline \emph{\scriptsize \{world, sports, business, technology\}} & 
        ``technology, business, sports, world, contracts, agreements, deals, real, estate, area'' &
        ``from the world of business, the world of technology, the world of sports, the world of politics,'' \texttt{...} \\
        \addlinespace[1.2ex]
        
        \textbf{Subj} \newline \emph{\scriptsize \{objective, subjective\}} & 
        ``subjective, objective, neutral, subjective, subjective, subjective, ...'' &
        ``:// Identify the tone of a text: formal, informal, objective, subjective
        The tone of the text is \texttt{...}'' \\
        \addlinespace[1.2ex]
        
        \textbf{TREC} \newline \emph{\scriptsize \{desc, entity, abbrev, human, loc, num\}} & 
        ``description, human, entity, location, name, title, date, number, time, amount'' &
        ``ainer The final answer is: \texttt{\$\textbackslash{}boxed\{6\}\$}assistant It seems like \texttt{...}'' \\
        
        \bottomrule
    \end{tabularx}
    \caption{Verbalizations produced by Classification DoD  translator versus InSPEcT, based on soft prompts trained on InSPEcT datasets. Repetitive and long sequences are truncated using (\texttt{...}).}
    \label{tab:InSPEcT_verbalization_comparison}
\end{table*}

\begin{table}[!htbp]
    \centering
    \small 
    \setlength{\tabcolsep}{3pt} 
    
    \begin{tabular}{l c c c c} 
        \toprule
        \textbf{Dataset} & \multicolumn{2}{c}{\textbf{Translator}} & \multicolumn{2}{c}{\textbf{InSPEcT}} \\
        \cmidrule(lr){2-3} \cmidrule(lr){4-5} 
        & \textbf{Recall} & \textbf{F1-Score} & \textbf{Recall} & \textbf{F1-Score} \\
        \midrule
        SST-2 & \textbf{1.0} & \textbf{1.0} & \textbf{1.0} & \textbf{1.0} \\
        SST-5 & \textbf{1.0} & \textbf{0.71} & 0.6 & 0.4 \\
        Subj & \textbf{1.0} & \textbf{0.8} & \textbf{1.0} & 0.5 \\
        AG-News & \textbf{1.0} & \textbf{0.57} & \textbf{1.0} & 0.44 \\
        TREC & \textbf{0.83} & \textbf{0.62} & 0.16 & 0.18 \\
        \bottomrule
    \end{tabular}
    \caption{Translator vs.~InSPEcT on InSPEcT datasets. The highest metrics for a dataset are bolded.}
    \label{tab:translator_vs_InSPEcT_quantitative}
\end{table}

Notably, our translator performs substantially better, in terms of  Recall, on SST-5 and TREC datasets where InSPEcT struggles due to a claimed dependency on high soft prompt task performance \cite{ramati2024elicitingtextualdescriptionsrepresentations}. In contrast, the translator's verbalization quality seems more robust to weak soft prompt task accuracy.
We further analyze the relationship between interpretability and task accuracy in Appendix~\ref{appendix:correlation_between_intepretability_and_task_accuracy}.

\textbf{Classification DoD}: As shown in Table~\ref{tab:translator_vs_InSPEcT_on_DoD_quantitative} (top), the translator verbalizations achieved an average F1-Score of 0.68 and an average Recall of 0.80. In contrast, we found that InSPEcT was completely unusable for this dataset, even after using the optimized layer pair \texttt{(27, 0)}, obtained using the training split of the DoD (see Appendix~\ref{appendix:InSPEcT_implementation_details}). Examples of verbalizations can be found in Table~\ref{tab:classification_DoD_verbalization_examples}.

\begin{table}[ht]
    \centering
    \small 
    \setlength{\tabcolsep}{3pt} 
    
    \begin{tabular}{l c c c c} 
        \toprule
        \textbf{DoD} & \multicolumn{2}{c}{\textbf{Translator}} & \multicolumn{2}{c}{\textbf{InSPEcT}} \\
        \cmidrule(lr){2-3} \cmidrule(lr){4-5} 
        & \textbf{Recall} & \textbf{F1-Score} & \textbf{Recall} & \textbf{F1-Score} \\
        \midrule
        Class  & 0.80 & 0.68 & 0 & 0 \\
        \midrule
        & \textbf{ROUGE-L} & \textbf{CosSim} & \textbf{ROUGE-L} & \textbf{CosSim} \\
        General  & 0.238 & 0.427 & 0.143 & 0.269 \\
        \bottomrule
    \end{tabular}
    \caption{Translator vs.~InSPEcT on Classification and General DoDs. ROUGE-L here corresponds to Prompt ROUGE-L. Results  are averaged over all constituent test datasets in each DoD.}
    \label{tab:translator_vs_InSPEcT_on_DoD_quantitative}
\end{table}


\begin{table*}[t]
    \centering
    \footnotesize 
    \setlength{\tabcolsep}{6pt} 
    
    \begin{tabularx}{\textwidth}{>{\raggedright\arraybackslash}p{2.5cm} 
                                 >{\raggedright\arraybackslash}X 
                                 >{\raggedright\arraybackslash}X} 
        \toprule
        \textbf{Dataset \& Classes} & \textbf{Our Translator Verbalization} & \textbf{InSPEcT Verbalization} \\
        \midrule
        
        \textbf{Instruments} \newline 
        \emph{\scriptsize \{drum, piano, violin, flute, guitar\}} &
        ``piano, violin, guitar, flute, drum, harp, cello, banjo'' &
        ``oga text: Identify the author of a given text: Shakespeare or Marlowe | \texttt{...} | Identify the author of a given text: Shakespeare or Marlowe | Identify'' \\
        \addlinespace[1.2ex]

        \textbf{Medical} \newline 
        \emph{\scriptsize \{analgesic, antifungal, antibiotic, antiviral, antipyretic\}} &
        ``antiviral, antibiotic, analgesic, antifungal, antiseptic, antibiotic'' &
        ``://www.grammarlytes.com
        The text is a link to a website that provides a free online grammar \texttt{...}'' \\

        
        \bottomrule
    \end{tabularx}
    \caption{Verbalizations produced using the classification DoD translator vs.~InSPEcT  on soft prompts from the Classification DoD.}
    \label{tab:classification_DoD_verbalization_examples}
\end{table*}

{\bf General DoD}:
\label{results:SuperNatural-DoD-translator}
As shown in Table~\ref{tab:translator_vs_InSPEcT_on_DoD_quantitative} (bottom), the translator achieved 0.238 Prompt ROUGE-L and 0.427 cosine similarity, averaged over all the testing samples in this DoD. It outperformed InSPEcT, which attained an average of 0.143 Prompt ROUGE-L and 0.269 cosine similarity, even after finding the optimized layer pair \texttt{(22, 1)} obtained using the training split of the DoD. Verbalization examples are shown in Table~\ref{fig:translator_and_inspect_supnat_example}.

\begin{table*}[ht] 
    \centering
    \footnotesize 
    \setlength{\tabcolsep}{6pt} 
    
    \begin{tabularx}{\textwidth}{>{\raggedright\arraybackslash}X >{\raggedright\arraybackslash}X >{\raggedright\arraybackslash}X} 
        \toprule
        \textbf{Ground-Truth Prompt} & \textbf{Translator (trained on General DoD)} & \textbf{InSPEcT \cite{ramati2024elicitingtextualdescriptionsrepresentations}} \\
        \midrule
        
        \textbf{task1391}: "In this task, you are given a question containing a blank (\_) and two options. You should pick the best option to answer the question. Please answer with ``A'' or ``B''." & 
        "Identify the correct answer from the given options using the provided context. Return the answer index: ``A'' or ``B''." &
        "ipel, use the same words to create a new answer. | Instruction: Use the same words to create a new answer. | ... | Instruction: Use the same words to create" \\
        \addlinespace[1.2ex] 
        \textbf{task392}: "In this task, you will be given two sentences separated by "because". You should decide whether the first sentence can be the result of the second sentence. If you can see a possible cause and effect relationship, answer with "plausible", otherwise answer with "not plausible". & 
        "Determine if the explanation for the given plot hole or inconsistency is plausible. If it makes sense and fits the context, label it as "plausible". Otherwise, label it as "not plausible". The explanation must address the specific plot hole or inconsistency." &
        ", Instruction: Use the first letter of the input to generate a word. The word should be a noun. | Instruction: Use the first letter of the input to generate a word. The word should be a noun. | Instruction: Use the first letter of the input to generate a word. The" \\
        \addlinespace[1.2ex] 
        
        \bottomrule
    \end{tabularx}
    \caption{Qualitative examples from General DoD, showing the translator's and InSPEcT's verbalizations of soft prompts for two tasks (task1391 and task392)}
    \label{fig:translator_and_inspect_supnat_example}
\end{table*}

\textbf{Comparison in terms of compute}: as alluded to in~\ref{section:related_works}, InSPEcT requires extensive hyperparameter optimization to find a "golden" layer-pair. For Llama-3.1-8b-Instruct, this involves an exhaustive search across 1,024 combinations to maximize metrics like Prompt ROUGE-L. This search is functionally a training phase, and is computationally expensive. Even on an NVIDIA A100 GPU, optimizing the layer-pair takes $\sim$25 minutes per task dataset, resulting in InSPEcT requiring $\sim$9 days of compute for the General DoD, and a prohibitive $\sim$95 days for the Classification DoD (we had to optimize on only a subset instead), and $\sim$2 hours for the InSPEcT datasets (See Appendix~\ref{appendix:InSPEcT_implementation_details}). 

For comparison, it takes $\sim$3 days to train a translator using the same base model and hardware (including soft prompt training, dataset compilation, and translator training). In the long-term, this is computationally much cheaper as a "train once, infer forever" open-source model that anyone can use off-the-shelf.


\subsubsection{Data Augmentation}
\label{results:data_augmentation}
The General DoD contains only 539 training examples, making it prone to overfitting if trained naively. Thus, we applied an augmentation procedure where we generated 10x paraphrases for each ground-truth hard prompt instruction. To validate the quality of paraphrases and ensure that they did not significantly alter the original semantics, we evaluated their performance (Task ROUGE-L) compared against the original instructions on the training set of General DoD. Both the original and paraphrased instructions had roughly similar performance ($\sim$0.2 Task ROUGE-L). 

In practice, we found that augmentation reduced overfitting and improved performance: the translator trained with augmentation achieved a Task ROUGE-L of 0.18, a statistically significant difference from the baseline which achieved Task ROUGE-L of 0.12 (paired t-test, $\alpha$ = 0.05) (see Appendix~\ref{appendix:data_augmentation_on_super_naturalinstructions_dod} for more details). We thus use the augmented translator for the remaining experiments in the paper.





\subsection{Task Relevance}
\label{results:relevance_and_utility_of_verbalizations}


To measure how informative and relevant a verbalization truly is to its task, we apply it as a prompt for inference. For each task, we sample 10 verbalizations from the translator, pass these to an LLM for inference on the training samples, select the verbalization with the highest training-set task performance, and report its performance on the test-set of each task. See Appendix~\ref{methods:post_training_experiments} for more details). We ran this experiment on the hold-out datasets of the General DoD using Llama-3.1-8B-Instruct as our LLM and compared against the following methods/baselines:

\begin{itemize}
  \setlength{\itemsep}{0pt}
  \setlength{\parskip}{0pt}
  \setlength{\parsep}{0pt}
  
  \item \textbf{InSPEcT}: We utilize InSPEcT to obtain verbalizations and use these as task prompt.

  \item \textbf{Baseline Prompt}: We utilize a generic prompt providing no information about the task: ``You will be given some input. Provide your best prediction of the output.''
  
  \item \textbf{Ground-Truth Task Description}: We use the ground truth task description as task prompt.
\end{itemize}

We utilized the same simple prompt template for all methods compared:
\begin{quote}
    \footnotesize \ttfamily
    \{Prompt\} \textbf{Input:} \{input\} \newline
    \textbf{Output:}
\end{quote}

\begin{figure}[!tb]
    \centering
    \includegraphics[width=1\linewidth]{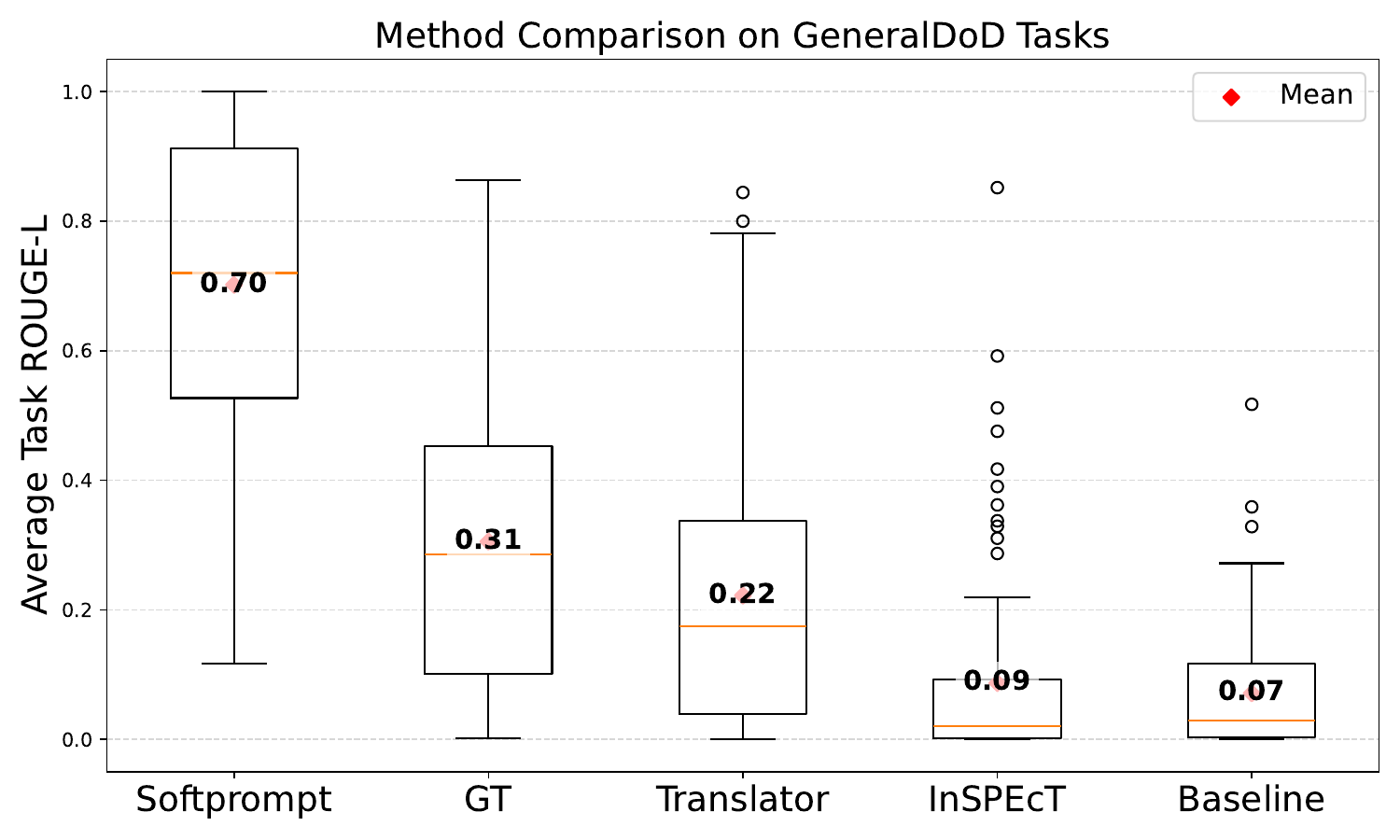}
    \caption{Comparison of translated prompts vs other strategies on Llama-3.1-8B-Instruct. Results are based on averaging task ROUGE-L across the hold-out set tasks in the General DoD.}
    \label{fig:fewshot_results_graph}
\end{figure}

Results (Figure \ref{fig:fewshot_results_graph}) show that the translator's verbalizations significantly outperform InSPEcT and the baseline generic prompt while approaching the performance of ground-truth task description. Using manual human validation, we found that the majority were partially related to the task; some verbalizations would get the task essentially exactly correct, while others missed by a wide margin. This is consistent with our quantitative comparison: around 84\% of the translator's verbalizations either matched or beat the baseline and 25\% matched or beat the ground-truth, suggesting the majority of verbalizations provided at least some useful information regarding the task. In contrast, around 47\% of InSPEcT's verbalizations matched or beat the baseline generic prompt.

Relative to the soft prompt's performance (0.7 avg Task ROUGE-L), the translator achieved 0.22 avg Task ROUGE-L, or roughly 32\% the performance of the soft prompt. While this is modest, it still represents a significant increase in how much information we are able to extract from soft prompts compared to InSPEcT's 13\% relative performance, which is the next closest method. Overall, we believe these results to be evidence suggesting that the translator is successfully extracting information from soft prompt embeddings and not just verbalizing plausible non-sense.

\begin{figure}[!htb]
    \centering
    \includegraphics[width=1\linewidth]{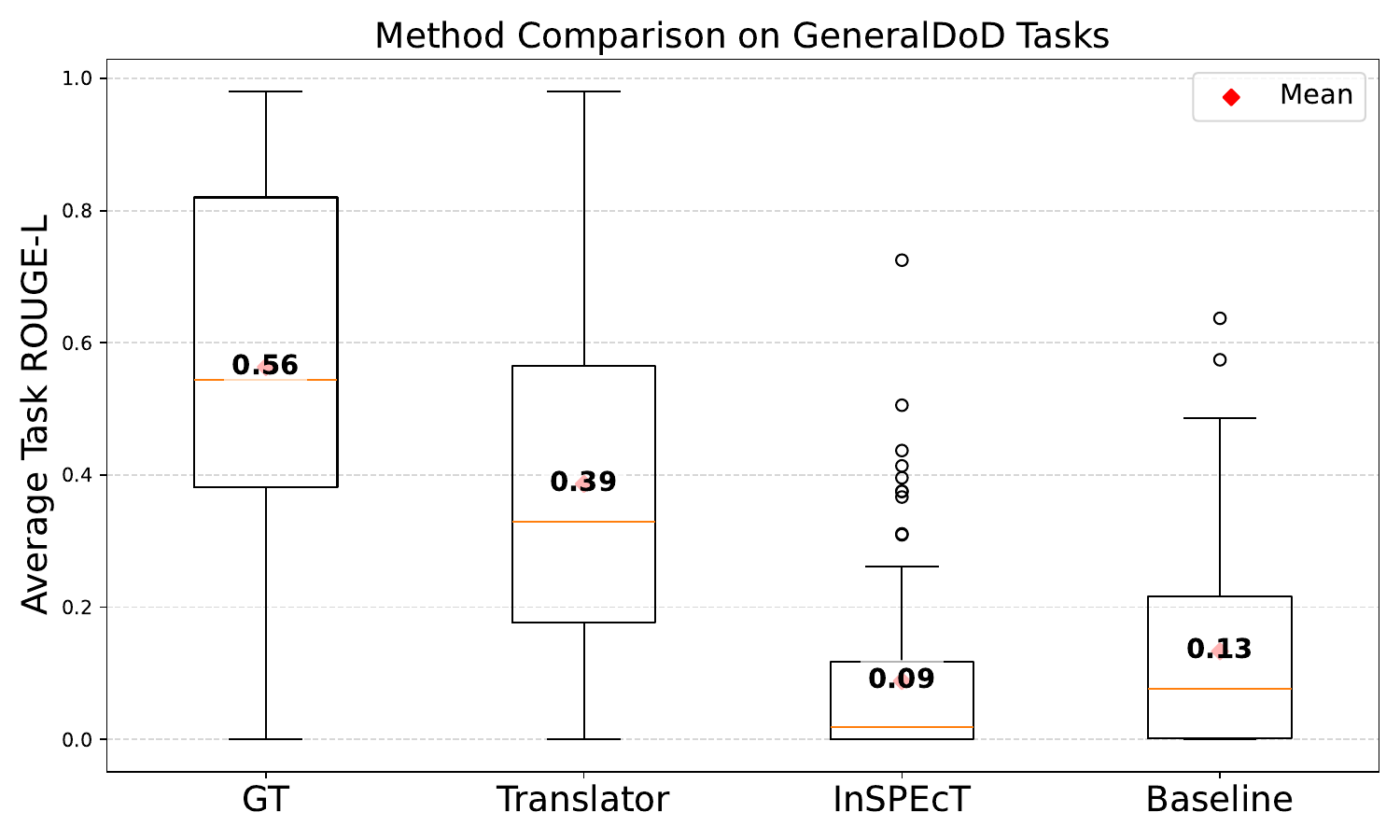}
    \caption{Comparison of translated prompts vs other strategies on a closed-source model (gpt-4o-mini). Results are based on averaging task ROUGE-L across the hold-out set tasks in the General DoD.}
    \label{fig:portability_graph}
\end{figure}

\textbf{Portability Test}: As an intuitive and quantitative way of measuring natural-language-ness of verbalizations, we test whether they are portable. Intuitively, natural language prompts are generally portable between models. Conversely, trying to port model-specific gibberish prompts to an unseen, closed-source model of a different family would likely fail. We experiment with applying verbalizations and other baselines on gpt-4o-mini and show there is no performance drop. All fluent prompts enjoy an increase, in contrast with InSPEcT which is now behind baseline prompting (See Figure~\ref{fig:portability_graph}).

\subsubsection{Out-of-Distribution Generalization}

As an out-of-distribution generalization test, we also evaluated General DoD translator on 7 classification datasets, which were never used to train or fine-tune it: Deceptive Spam Corpus \citep{li-etal-2014-towards}, Human vs AI Generated Essays \citep{kaushal2023humanvsai}, and the five InSPEcT datasets. We then analyzed the verbalizations (see Table~\ref{tab:OOD_examples}). Overall, we see that the translator is able to verbalize the classes in six out of seven datasets, averaging a class rate of 0.78.

\begin{table*}[t] 
    \centering
    \small 
    \setlength{\tabcolsep}{6pt} 
    
    \begin{tabularx}{\textwidth}{@{} >{\raggedright\arraybackslash}p{3.5cm} >{\raggedright\arraybackslash}X @{}} 
        \toprule
        \textbf{Dataset} & \textbf{Verbalizations} \\
        \midrule
        
        \textbf{SST-2} & 
        ``Classify each movie review as "\textbf{positive}" or "\textbf{negative}" based on its content. Use the exact tags: "\textbf{positive}" or "\textbf{negative}".'' \\
        \addlinespace[1ex]
        
        \textbf{Subj} & 
        ``Classify the Amazon food product review into one of these categories: "\textbf{subjective}" or "\textbf{objective.}" A review is "\textbf{subjective}" if it expresses personal opinions, feelings, or experiences. A review is "\textbf{objective}" if it can be answered with a yes/no question and includes product details like price, weight, or ingredients.'' \\
        \addlinespace[1ex]
        
        \textbf{AG-News} & 
        ``Classify the news article into one of these categories: "\textbf{world}," "\textbf{sports}," "\textbf{business}," "\textbf{sci-tech}," "entertainment," or "health." Use the exact tags specified for each category. Ensure each article is classified into only one category.'' \\
        \addlinespace[1ex]
        
        \textbf{TREC} & 
        ``Identify the question's category: \textbf{entity}, \textbf{location}, or \textbf{description}. For \textbf{entity}, output the object's name. For \textbf{location}, output the city's name. For \textbf{description}, output the phrase describing the object. Preserve exact sentence counts and structural relationships.'' \\
        \addlinespace[1ex]

        \textbf{SST-5} & 
        ``Classify the given sentence as "INVITATION" if it is an invitation to join a community or a group. Classify it as "RETWEET" if it is a call for retweeting a tweet. Otherwise, classify it as "INFO".'' \\
        
        \textbf{Deceptive Spam Corpus} & 
        ``Classify the given English sentence as either "\textbf{truthful}" or "\textbf{deceptive}" based on whether it is intentionally meant to deceive the receiver. A sentence is considered \textbf{deceptive} if it manipulates the receiver's beliefs about the object, person, or event. The sentence may contain a single sentence or multiple sentences'' \\
        \addlinespace[1ex]
        
        \textbf{Human vs AI} & 
        ``Classify the article as "\textbf{Human}" or "\textbf{Machine}" based on whether it was written by a human or generated by a machine. Output: "\textbf{Human}" or "\textbf{Machine}"'' \\
        \addlinespace[1ex]
        
        \bottomrule
    \end{tabularx}
    \caption{Qualitative examples of verbalizations produced by the General Translator on out-of-distribution classification soft prompts. The true class labels in the dataset are bolded.}
    \label{tab:OOD_examples}
\end{table*}









\section{Discussion}
In our experiments, we found that the translators consistently outperformed the InSPEcT method and other baselines on the InSPEcT datasets, the Classification DoDs, and General DoDs, both qualitatively and quantitatively. The Classification translator's verbalizations displayed strong performance at extracting class label information, while the General translator's verbalizations were mostly good. Based on manual human evaluation, both translators produced fluent and human-readable text. On out-of-distribution evaluation datasets, the General translator was able to produce accurate verbalizations on six out of seven datasets tested. 

We think this suggests a promising proof-of-concept for a method. It is conceivable that in the future, a more powerful translator could be used as a tool to extract information from the learned parameters of a soft prompt and used on downstream applications such as an interpretability tool to learn about how models are solving tasks, or as a way to obtain discrete, portable hard prompts for inference. Speculatively, perhaps a similar approach could even be extended in the future to other machine learning techniques such as trying to verbalize information encoded within the parameters of a LoRA adapter into natural language.



\section{Conclusion and Future Work}
\label{conclusions}



In this work, we investigated the idea of training a model to decode information stored in the parameters of a soft prompt into natural language. We demonstrate a proof-of-concept translator model capable of outputting fluent natural language task descriptions from a soft prompt's learnable embeddings that exceeds current methods and suggests promise.

In our experiments, we trained our translators using textual descriptions of the task as the target verbalization due to data limitations. However, a soft prompt likely encodes more complicated procedures not fully contained by a description of the task alone. In practice, this means the current General translator is trained to extract information about `what' the task, but not information such as `how' to solve (since that was never present in its training set). For instance, it may verbalize information about ``This is a classification task...'' but it does not verbalize information about what features are to be used to solve the task. This is fine for a proof-of-concept, but requires more work to become a practical interpretability method.

Conceivably, in the future this could be addressed by creating a dataset tailored for translator training. Intuitively, solving a task involves solving a number of sub-tasks. For instance, a sentiment analysis task might contain sub-tasks like: `count how many times x negative word appears' Verbalizing the `how' to solve such tasks just involves verbalizing `what' all the sub-tasks are. Because the translator already understands to a degree how to verbalize the `what' part, in some ways the hard part has already been tackled. In the future, we could consider engineering a new DoD where each task (target hard prompt) is a composition of sub-tasks which the translator recognizes. The training objective would then be to verbalize all these sub-tasks, thus in principle, teaching it how to output more detailed verbalizations.

\section{Limitations}

We did not explore in this work whether our soft prompt translator model can work with soft prompts trained on a different base model, though given that soft prompts are not normally transferrable between models, it is safe to assume this is not the case. This means that unless the soft prompt is already trained using the same base model as ours (Llama-3.1-8B-Instruct), a new translator model would need to be trained. This is inconvenient, but not prohibitive given that it only has to be done once per model. In practice, this just looks like fine-tuning translators for the commonly-used open-source language models, then releasing it for others to use.

Experimentally, we also were limited to just a single closed source model. It is not immediately apparent that this method would fail on other open source language models, but this is not verified either. So in the mean time, we present these results as a model-specific proof-of-concept. 


\section*{Acknowledgments}

Parts of this work utilized AI for coding assistance for miscellaneous tasks such as visualizations and implementations of certain methods, etc. All implementations were checked and tested by humans. No AI assistance was used for the ideation or writing of the paper itself.



\nocite{*}
\bibliography{custom}

\appendix

\section{Additional details}
\label{sec:appendix}

\subsection{Code Repository, Dataset, and Translator Weights}
\label{appendix:src_code}
All code related to our experiments can be found at \href{https://anonymous.4open.science/r/softprompt_experiments-B072}{this} anonymous remote repository. We also release the DoDs and trained translators publicly for further research in the soft prompt interpretability domain:
\begin{itemize}
    \setlength{\itemsep}{0pt}
    \setlength{\parskip}{0pt}
    \setlength{\parsep}{0pt}
    \item \href{https://anonymous-hf.com/a/oei056pd14xk/}{Classification-DoD}
    \item \href{https://anonymous-hf.com/a/azsena271em9/}{General-DoD-10x}
    \item \href{https://anonymous-hf.com/a/l4tpf6m2uo6b/}{General-DoD-translator}
    \item \href{https://anonymous-hf.com/a/3v0phfo40r2l/}{General-DoD-translator-10x}
    \item \href{https://anonymous-hf.com/a/yhzzp7rnpal8/}{Classification-DoD-translator}
\end{itemize}

\subsection{Synthetic dataset creation details}
\label{appendix:synthetic_dataset_creation_details}
For Classification-DoD generation, we utilized two separate vLLM \cite{vllm_kwon2023efficient} pipelines for related keyword generation and sentence generation. 

\textbf{Keyword Generation}: To choose keywords for a mini-dataset, we used a vLLM pipeline which sampled 3 random (nouns or adjectives) from the Brown Corpus \cite{browncorpus}, and primed an LLM with the sampled random keywords as ``abstract inspiration'' to generate distinct words as classes which are related by concepts, industries, themes, etc. For each dataset, we generated 5 keywords as the dataset's classes. We used Mistral-Small-3.1-24B-Instruct-2503 as the LLM for keyword generation.

\textbf{Sentence Generation}: Using chosen keywords per mini-dataset, we generated 
100 sentences per keyword (500 sentences per dataset), using Mistral-Small-3.1-24B-Instruct-2503. This dataset generation pipeline uses the vLLM library to efficiently batch requests together for higher generation throughput.  Each mini-dataset was stored in a single SQLite3 database file, in a relational database manner on 3 tables, for fast and easy retrieval during the soft prompt training phase. For each mini-dataset, we store the generated sentences with their ground-truth class label, which was used to generate the sentences. Due to the constraint that the classes selected must be related to each other, we had instances where some mini-datasets contained classes which are very difficult to differentiate such as: \texttt{revive, stimulate, inspire, encourage, motivate}, which are generated from using the abstract word \texttt{energize}. Due to this, we observed many soft prompts with task accuracy below 50\%.

\subsection{InSPEcT experiment details}
\label{appendix:InSPEcT_implementation_details}
We trained soft prompts of 20 tokens in length on the SST-2, SST-5, AG-News, Subj, and TREC datasets, according to the hyperparameter configurations mentioned in Table 4 of the InSPEcT Paper \cite{ramati2024elicitingtextualdescriptionsrepresentations}. We utilized the Meta-Llama-3.1-8B-Instruct model as the base LLM for soft prompt training. For the patching experiments, we utilized the same LLM, which has 32 decoder layers. For determining a specific layer-pair hyperparameter: there are 1024 combinations (32 * 32, which excludes the last decoder layer and includes the embedding layer). The existing code for InSPEcT paper uses limited layer-pair combinations \texttt{(source layer, target layer)}, though our experiments used all layer-pair combinations, as the InSPEcT paper and code repository does not mention the "golden" layer pairs to use. 

We also applied InSPEcT to our Classification DoD and General DoD, by optimizing for the layer-pair hyperparameter on the training split of each DoD. For this, we performed a grid search over all 1,024 layer pair combinations and computed an evaluation metric. We then picked the layer-pair which provided the best evaluation metric average value across all mini-datasets in the DoD. For optimization, we used the following evaluation metrics: average of F1-Score and Class Rate for Classification DoD and Prompt ROUGE-L and cosine similarity for General DoD. Since we established in section~\ref{results:comparing_ground_truth_alignment} that InSPEcT on Classifcation and General DoDs will be expensive, we reduced the training set size to just first 50 examples, to optimize the layer-pair hyperparameter in a scalable way.

Our implementation of InSPEcT is forked from the \href{https://github.com/danaramati1/InSPEcT}{original InSPEcT GitHub repository}. For InSPEcT datasets, we used the following dataset repositories on Hugging Face:

\begin{enumerate}
    \setlength{\itemsep}{0pt}
    \setlength{\parskip}{0pt}
    \setlength{\parsep}{0pt}
    \item \textbf{SST-2}: \texttt{stanfordnlp/sst2}
    \item \textbf{SST-5}: \texttt{SetFit/sst5}
    \item \textbf{AG-News}: \texttt{fancyzhx/ag\_news}
    \item \textbf{Subj}: \texttt{SetFit/subj}
    \item \textbf{TREC}: \texttt{SetFit/TREC-QC}
\end{enumerate}

\subsection{Soft Prompt Training Implementation Details}
\label{appendix:soft_prompt_training_implementation_details}
We used the PEFT library \cite{peft} for training soft prompts for InSPEcT, whereas we used our own custom implementation (see \href{https://anonymous.4open.science/r/softprompt_experiments-B072/src/softprompt_experiments/models/softprompt.py}{Custom Soft Prompt}) for training soft prompts on DoDs. We also reversed the implementations for each method (for example: trained InSPEcT soft prompts using custom implementation) and got worse results for both. Therefore, we report the best results above for each method, across all implementations.

\subsection{Soft prompt Translator training details}
\label{appendix:soft_prompt_translator_training_details}
All methods described in this section use soft prompt embeddings as the input sequence and hard prompts as the output sequence for training using Maximum Likelihood Estimation. As described in our paper, we model soft prompt translation as a sequence to sequence, text-generation task where we pass in soft prompt embeddings as input and calculate supervised, next-token loss over the target hard prompt output sequence. We fine-tune Llama-3.1-8B-Instruct using LoRA \cite{hu2021loralowrankadaptationlarge}. For LoRA implementation, we used the PEFT library with the following LoRA configuration:

\begin{lstlisting}[language=Python, caption={LoRA configuration used with the PEFT library.}, label={code_block:lora_config}]
lora_config = LoraConfig(
    r = LORA_RANK, 
    lora_alpha = 2 * LORA_RANK,
    target_modules = ["q_proj", "v_proj", "k_proj", "o_proj", "gate_proj", "up_proj", "down_proj"],
    lora_dropout = LORA_DROPOUT,
    bias = "none",
    task_type = TaskType.CAUSAL_LM
)
\end{lstlisting}

\subsection{Preliminary Classification DoD vs Classification DoD}
\label{appendix:prelim_classification_Dod_vs_classification_DoD}

An earlier iteration of our Classification DoD --- Preliminary Classification DoD ---- lacked semantic relationship constraint between keywords within a mini-dataset. 
In terms of soft prompts trained on InSPEcT datasets, translator trained on Classification DoD outperformed translator trained on Preliminary Classification DoD, not only in terms of Class Rate and F1-score, but also in terms of quality of the verbalizations produced and out-of-distribution generalization. This can be attributed to the constraint placed on the selection of keywords within each mini dataset in Classification DoD (which we discussed in section~\ref{appendix:synthetic_dataset_creation_details}),  which allowed the mini datasets, and hence the respective trained soft prompts, to be more real-world-like. The comparison of Classification DoD translator and Preliminary Classification DoD can be found in Table~\ref{tab:DoD1_vs_DoD2_translator_on_InSPEcT_datasets}. Hence for further experiments, we use Classification DoD translator.

\begin{table*}[!htb]
    \centering
    \small 
    \begin{tabular}{>{\raggedright\arraybackslash}p{2cm} 
                    >{\raggedright\arraybackslash}p{2cm}
                    >{\raggedright\arraybackslash}p{2cm}
                    >{\raggedright\arraybackslash}p{2cm}
                    >{\raggedright\arraybackslash}p{2cm}} 
        \toprule
        \textbf{Dataset} &
        \textbf{Class Rate (Preliminary Classification DoD translator)} &
        \textbf{F1-Score (Preliminary Classification-DoD translator)} &
        \textbf{Class Rate (Classification DoD translator)} &
        \textbf{F1-Score (Classification DoD translator)} \\
        \midrule

        \textbf{SST-2} & 
        1.0 & 0.33 & 
        \textbf{1.0} & \textbf{1.0} \\
        \addlinespace

        \textbf{SST-5} & 
        0.4 & 0.4 &
        \textbf{1.0} & \textbf{0.71} \\
        \addlinespace 

        \textbf{AG-News} & 
        0.75 & 0.46 &
        \textbf{1.0} & \textbf{0.57 }\\
        \addlinespace 

        \textbf{Subj} & 
        1.0 & 0.57 &
        \textbf{1.0 }& \textbf{0.8 }\\
        \addlinespace 

        \textbf{TREC} & 
        0.33 & 0.27 &
        \textbf{0.83} & \textbf{0.62} \\
        \addlinespace 
        
        \bottomrule
    \end{tabular}
    \caption{Comparison of Classification DoD translator and Preliminary Classification DoD translator on soft prompts trained on the InSPEcT datasets (out-of-distribution for both)}
    \label{tab:DoD1_vs_DoD2_translator_on_InSPEcT_datasets}
\end{table*}

\subsection{Preprocessing SuperNatural Dataset}
\label{appendix:preprocessing_supernatural_instructions_dataset}
Since the original SuperNatural dataset is in a \href{https://github.com/allenai/natural-instructions}{GitHub repository}, comprising of multiple JSON files, we decided to parse this into an easy-to-use dataset on Hugging Face. We performed some data analysis and decided to further filter the dataset for training soft prompts and translators on it. Based on our data analysis, we performed the following preprocessing:
\begin{enumerate}
    \setlength{\itemsep}{0pt}
    \setlength{\parskip}{0pt}
    \setlength{\parsep}{0pt}
    \item Filter out any instance sequences (input + output) longer than 400 tokens, as we found that there were certain instances where the sequence length was over 200,000 tokens, with P99 of 1312 and P90 of 337. To keep at least 90\% of the instance data, we picked 400 tokens as the threshold for this filtration.
    
    \item Filter out any tasks with less than 500 instances, as we found that some tasks had a minimum of 29 instances only, which would not be sufficient for training soft prompts. Hence, we decided to use 500 instances as a minimum threshold for selecting tasks for future experiments.

    \item Added a column for total tokens of input + output sequences, calculated using tokenizer of Llama-3.1-8B-Instruct language model, as we dynamically use it later during soft prompt training, for data collation.

    \item Created instances of (input, output[$i$]) pairs (unwinding), if the output is an array of multiple values for a given input, for all values of $i$ in the output array. 
\end{enumerate}

In the end, we were left with 539 tasks for training split and 96 tasks for testing split.

\subsection{Data Augmentation on General DoD}
\label{appendix:data_augmentation_on_super_naturalinstructions_dod}

In order to test our hypothesis of whether the translator quality improves with more training data, we decided to perform data augmentation to the training split of General DoD. We used a vLLM pipeline, using the Mistral-Small-3.1-24B-Instruct-2503 LLM, to generate 10 paraphrases of the original instructions using the following system prompt:

\begin{lstlisting}[language=Python, caption={System Prompt for generating paraphrases with reduced number of tokens}, label={system_prompt_for_paraphrasals}]
system_prompt = (
        "You are an expert at paraphrasing instructions. "
        "Your task is to rephrase the following instruction using different words and sentence structures while keeping the exact same meaning and level of detail. "
        "CRITICAL: You MUST explicitly preserve all specific classes, exact tags, labels, output formats, and special syntax constraints. "
        "CRITICAL: Do NOT alter or remove specific mappings between concepts (e.g., specifying which sentence is the premise, exact sentence counts, positional logic, or structural relationships). "
        "Output ONLY the paraphrased instruction text and absolutely nothing else. Do NOT include conversational filler like 'Here is the paraphrased version'."
    )
\end{lstlisting}

 We generated a new training split with 10x augmentation applied and this is used to train \texttt{translator10x}, described in section~\ref{results:data_augmentation}. 

 To validate if the generated paraphrased ground-truth task description does not significantly alter the original semantics, we performed the following test: compared task accuracy (Task ROUGE-L) of using ground-truth task description and 1 randomly sampled paraphrase of it. We found both values to be 0.236 and 0.269, respectively, suggesting that they are similar in terms of task relevance.

\subsection{Correlation between Interpretability and Task Accuracy}
\label{appendix:correlation_between_intepretability_and_task_accuracy}
While testing Classification DoD translator on the testing split of Classification DoD, we observed an interesting phenomenon: the accuracy of the soft prompts does not need to be high in order to get good verbalizations out of them. For example, most verbalizations of the soft prompts from the testing split of Classification DoD are commendable in terms of Recall and F1-score, despite having low task performance (< 50\%). This is contrary to the idea proposed in the InSPEcT paper, where they claim a correlation between quality of verbalizations and the task accuracy of the soft prompts. Some verbalizations for soft prompts with low task accuracy can be found in Table~\ref{tab:Classification_DoD_test_verbalization_examples}.

We hypothesize that while the soft prompt applied on the LLM may achieve poor performance, this might not just be because they encode irrelevant task instructions; another valid possibility might be that the soft prompt does encode relevant task instructions and our small base model of only 8 billion parameters (Llama-3.1-8B-Instruct) may be failing to execute them.


\begin{table*}[t]
    \centering
    \small 
    \begin{tabular}{>{\raggedright\arraybackslash}p{4cm} 
                    >{\raggedright\arraybackslash}p{4cm}
                    >{\raggedright\arraybackslash}p{2cm}
                    >{\raggedright\arraybackslash}p{2cm}
                    >{\raggedright\arraybackslash}p{2cm}} 
        \toprule
        \textbf{Hard Prompt} & 
        \textbf{Verbalized Soft Prompt} &
        \textbf{Soft Prompt Accuracy} &
        \textbf{Recall} &
        \textbf{F1-score} \\
        \midrule

        \textbf{anchored, permanent, fixed, immovable, stationary} & 
        stationary, fixed, anchored, permanent, immovable, sailing, moving, floating, transient &
        51.67\% & 1.0 & 0.71 \\
        \addlinespace 

        \textbf{abundant, profitable, bountiful, prosperous, plentiful} & 
        abundant, plentiful, fruitful, bountiful, prosperous, rich, successful, booming &
        75.0\% & 0.8 & 0.62 \\
        \addlinespace 

        \textbf{revival, rejuvenation, restoration, recuperation, regeneration} & 
        restoration, regeneration, revival, rejuvenation, restoration, recuperation, revival, regeneration, &
        57.45\% & 1.0 & 1.0 \\
        \addlinespace 

        \textbf{pity, empathy, compassion, mercy, sorrow} & 
        compassion, sorrow, mercy, pity, empathy &
        41\% & 1.0 & 1.0 \\
        \addlinespace 

        \textbf{gps, lidar, braille, sonar, radar} & 
        model, google, braille, echolocation, sonar &
        93.13\% & 0.6 & 0.50 \\
        \addlinespace 
        
        \bottomrule
    \end{tabular}
    \caption{Qualitative examples of verbalizations produced using Classification-DoD-translator on the soft prompts from the testing split of Classification-DoD.}
    \label{tab:Classification_DoD_test_verbalization_examples}
\end{table*}

\subsection{Post-Training Optimization}
\label{methods:post_training_experiments}
To further improve the verbalization quality of the trained General DoD translator, we decided to optimize using task accuracy (Task ROUGE-L), as a `scoring' metric, on the training \textit{(input, output)} sequences for each task in the General DoD (same sequences used to train the soft prompt). We tried the following method for all tasks in General DoD: start by translating a given soft prompt, using greedy decoding (temperature = 0), to obtain a baseline hard prompt, $z_G$, and then calculate its task accuracy (also called score $s(z_G)$), using the same method described in section~\ref{results:relevance_and_utility_of_verbalizations} and \texttt{meta-llama/Llama-3.1-8B-Instruct} as the task LLM. We then generate $N$ alternative hard prompts ($z_1, \dots, z_N$) for the same soft prompt via temperature sampling (temperature > 1). If any sampled prompt $z_i$ outperforms the baseline such that $s(z_i) > s(z_G)$, we designate it as $z_W$, which we call the `optimized' translation of the soft prompt. If no $z_i$ satisfies this condition, we incrementally increase the sampling temperature and repeat the process until a maximum temperature threshold is reached. This method aims to extract the optimal hard prompt from the translator's learned distribution; we refer to this approach as \texttt{translator++}. For our sampling procedure, we set top-$p$ to 0.9, the initial temperature to 0.5 (with step increments of 0.25), and the maximum temperature to 1.5.

We also tried Direct Preference Optimization (DPO) \cite{rafailov2024directpreferenceoptimizationlanguage} as a post-training alignment technique. We constructed a preference dataset by designating $z_W$ as the preferred output. For the dispreferred output ($z_L$), we selected the translation $z_i$ that achieved the lowest score among all sampled generations (including $z_G$).

DPO implicitly defines a reward function 
\begin{equation}
    r_\theta(y \mid z') = \beta \log \frac{\pi_\theta(y \mid z')}{\pi_{\text{ref}}(y \mid z')}
\end{equation}
where $\beta$ controls the weight of the implicit KL penalty. Using this, we optimize the following objective:
\begin{equation}
\begin{split}
    \mathcal{L} &= -\mathbb{E}_{\mathcal{D}} \Big[ \log \sigma \big( r_\theta(z_W \mid z') - r_\theta(z_L \mid z') \big) \\
    &\quad - \log \pi_\theta(z_W \mid z')\Big]
\end{split}
\label{eq:dpo_loss}
\end{equation}
where the expectation is taken over triplets $(z', z_W, z_L)$ from our preference dataset $\mathcal{D}$, $\pi_\theta$ is the policy model, $\pi_{\text{ref}}$ is the frozen reference model, and $\sigma$ is the logistic sigmoid function. We also included an additional Supervised Fine-Tuning (SFT) loss term ($\log \pi_\theta(z_W \mid z')$), for regularization in equation~\ref{eq:dpo_loss}. We tried training using DPO for two rounds: 

\begin{itemize}
  \setlength{\itemsep}{0pt}
  \setlength{\parskip}{0pt}
  \setlength{\parsep}{0pt}
  
  \item \textbf{Round 1}: We generated a preference dataset using $\pi_{\text{ref}}$ model as \texttt{translator10x}. We then trained using $\pi_\theta$ as \texttt{translator10x} and obtained \texttt{DPO-translator-1}.

  \item \textbf{Round 2}: Using \texttt{DPO-translator-1} as $\pi_{\text{ref}}$, we generated another preference dataset and use this dataset to train \texttt{DPO-translator-2}, using $\pi_{\text{ref}}$ as \texttt{DPO-translator-1}.

\end{itemize}

Results are presented in Table~\ref{tab:DPO_results}, in terms of Task Accuracy computed using \texttt{meta-llama/Llama-3.1-8B-Instruct} as the Task LLM. We also experimented with different $\beta$ values. We use \texttt{translator++} and \texttt{translator10x} as baselines for comparison. Although DPO Round 2 shows the best performance in terms of Task ROUGE-L, the performance improvements remain marginal over baselines. Refining the DPO method and scaling the preference datasets are thus left for future work.

\begin{table}[ht]
    \centering
    \small 
    \setlength{\tabcolsep}{4pt} 
    
    \begin{tabular}{@{} c c c c c c @{}} 
        \toprule
        \multicolumn{2}{c}{\textbf{DPO Round 1}} & \multicolumn{2}{c}{\textbf{DPO Round 2}} & \multicolumn{2}{c}{\textbf{Baselines}} \\
        \cmidrule(lr){1-2} \cmidrule(lr){3-4} \cmidrule(l){5-6}
        $\beta = 0.1$ & $\beta = 1.0$ & $\beta = 0.1$ & $\beta = 1.0$ & \textbf{Trans++} & \textbf{Trans10x} \\
        \midrule
        0.22 & 0.22 & 0.21 & \textbf{0.23} & 0.22 & 0.18 \\ 
        \bottomrule
    \end{tabular}
    \caption{Comparison of Task ROUGE-L of Round 1 and Round 2 DPO experiments across different $\beta$ values against baseline translators.}
    \label{tab:DPO_results}
\end{table}

\end{document}